\documentclass[10pt,twocolumn,letterpaper]{article}

\usepackage{iccv}
\usepackage{amsmath,graphicx}
\usepackage{times}
\usepackage{epsfig}
\usepackage{amssymb}
\usepackage{nicefrac}
\usepackage{epstopdf}
\usepackage{booktabs}
\usepackage{multirow}
\usepackage{subfigure}
\usepackage{dblfloatfix}
\usepackage[linesnumbered,ruled]{algorithm2e}
\usepackage{algpseudocode}
\usepackage{bm}
\usepackage{dsfont}
\DeclareMathAlphabet\mathbfcal{OMS}{cmsy}{b}{n}
\usepackage[pagebackref=true,breaklinks=true,letterpaper=true,colorlinks,bookmarks=false]{hyperref}
\usepackage{array}
\newcolumntype{P}[1]{>{\centering\arraybackslash}p{#1}}

\iccvfinalcopy 

\def\iccvPaperID{10} 

\ificcvfinal\fi
\begin{document}

\title{Adaptive SVM+: Learning with Privileged Information for Domain Adaptation}

\author{Nikolaos Sarafianos \qquad Michalis Vrigkas \qquad Ioannis A. Kakadiaris\\
	Computational Biomedicine Lab, University of Houston\\
	\small{\texttt{\{nsarafia, mvrigkas, ikakadia\}@central.uh.edu}}
}

	
	\maketitle   
\thispagestyle{empty}

\begin{abstract}    
	Incorporating additional knowledge in the learning process can be beneficial for several computer vision and machine learning tasks. Whether privileged information originates from a source domain that is adapted to a target domain, or as additional features available at training time only, using such privileged (\ie, auxiliary) information is of high importance as it improves the recognition performance and generalization. However, both primary and privileged information are rarely derived from the same distribution, which poses an additional challenge to the recognition task. To address these challenges, we present a novel learning paradigm that leverages privileged information in a domain adaptation setup to perform visual recognition tasks. The proposed framework, named Adaptive SVM+, combines the advantages of both the learning using privileged information (LUPI) paradigm and the domain adaptation framework, which are naturally embedded in the objective function of a regular SVM. We demonstrate the effectiveness of our approach on the publicly available Animals with Attributes and INTERACT datasets and report state-of-the-art results in both of them.
\end{abstract}

\section{Introduction}\label{sec:intro}
When Vapnik and Vashist introduced the learning using privileged information (LUPI) framework \cite{vapnik2009new}, they drew inspiration from human learning. They observed how significant the role of an intelligent teacher was in the learning process of a student, and proposed a machine learning paradigm to imitate this process. Distilling knowledge in the learning process can take many forms, which impact the training stage in different ways. Privileged information can appear in the form of additional features available only at training time \cite{Pechyony_2011_15083}, in the form of a curriculum learning strategy \cite{bengio2009curriculum, jiang2014self} (\ie, presenting easier examples before more complicated), or by transferring feature representations to other domains \cite{bousmalis2016domain, sun2015subspace} by incorporating the adaptation to a new domain in the learning process \cite{bengio2012deep, long2016deep, yang2007cross}. However, what is considered as privileged information, how it can be incorporated in the learning process, and in what form, depends on the task, the available features, and the learning scheme (supervised, or semi-supervised). 

\begin{figure*}[t] 
	\centering
	\includegraphics[width=0.77\textwidth]{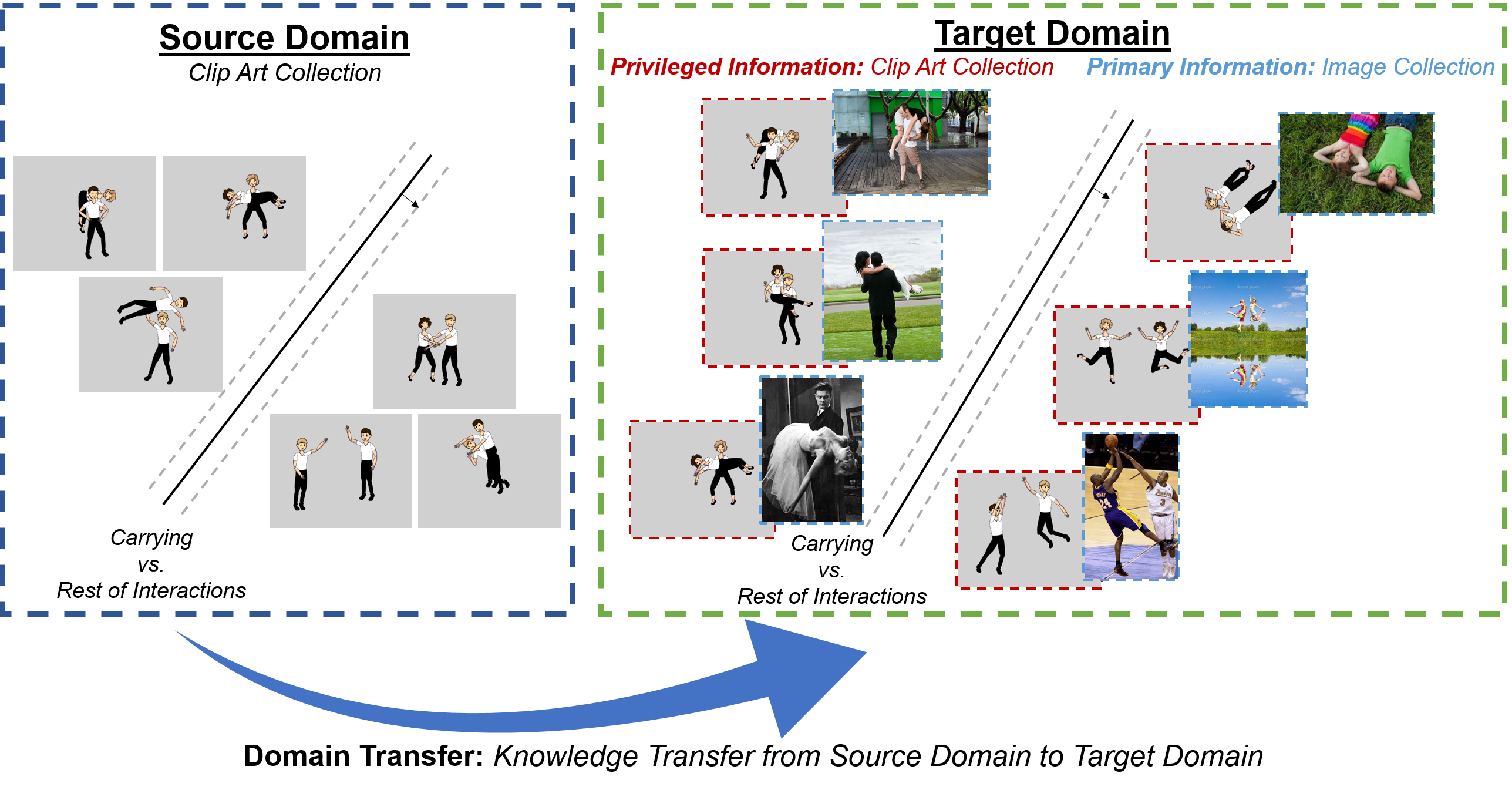}
	\caption{Can we leverage privileged information in a domain adaptation setup? Wouldn't it be great if we could find a way to leverage information from a source domain and at the same time employ privileged information in the target domain? Our proposed approach aspires to combine the advantages of domain transfer and the learning using privileged information paradigm to solve visual recognition tasks.} 
	\label{fig:SchemeParadigm}
\end{figure*}

The scope of this work is to train a better classifier and not to perform an end-to-end learning process to obtain better features. The proposed scheme is general and can be applied to any type of features (\eg, features extracted from the last fully-connected layer of the VGG network \cite{simonyan2014very}). We tested our method in object recognition and human interaction classification tasks, using as privileged information visual attributes and clip art illustrations respectively, and human annotation scores (easy/hard) to obtain the different domains. An illustrative example of our method is depicted in Figure~\ref{fig:SchemeParadigm}. 

In this work, we aspire to exploit privileged information in a two-fold manner: first as additional information that is available only during training but not at testing time, and second, by learning representations in a source domain and transferring this information to a target domain. We combine the advantages of the LUPI paradigm \cite{vapnik2009new} and domain adaptation as proposed by Yang \etal~\cite{yang2007cross} and introduce Adaptive SVM+; a new learning scheme that incorporates privileged information (SVM+) and knowledge transferred from a source domain to a target domain (Adaptive SVM) in the objective function to improve performance and generalization.

\section{Related Work and Prior Knowledge}\label{sec:rwork}
\noindent\textbf{Privileged Information}: The idea of leveraging additional information during the learning phase is not a new concept as it has previously been addressed in the literature in many contexts. The choice of different types of privileged information in the context of object recognition implemented in a max-margin scheme was proposed by Sharmanska \etal~\cite{sharmanska2013learning}. Furthermore, Wang and Ji \cite{wang2015classifier} proposed two different loss functions that exploit privileged information and can be used with any classifier. The first model encoded privileged information as an additional feature during training, while the second approach considered that privileged information can be represented as secondary labels. An interesting method that discusses the auxiliary view (\ie, auxiliary information) from an information theoretic perspective was introduced by Motiian \etal~\cite{motiian2016information} and was also extended to unsupervised domain transfer \cite{motiian2016informationeccv}. Lapin \textit{et al.}~\cite{Lapin_2014_16384} related the privileged information framework to the importance of sample weighting and showed that prior knowledge can be encoded using weights in a regular SVM. Recently, the LUPI paradigm has been employed with applications on gender classification facial expression recognition as well as age and height estimation \cite{kakadiaris2016body, sarafianos2016predicting, vrigkas2016exploiting, Wang_2015_16756}. 

	\noindent\textbf{Knowledge Distillation and Curriculum Learning}: Transfer learning seeks to leverage the knowledge obtained while learning some tasks and applying it to new unseen, and possibly unrelated, tasks. It has been applied with great success in applications ranging from cross-domain setups \cite{duan2012domain, gupta2016cross, yang2007cross}, to facial action unit detection in transductive learning setup \cite{chu2013selective},  to deep neural networks \cite{hinton2015distilling, zhang2016real}. Lopez-Paz \etal~\cite{LopSchBotVap16} introduced generalized distillation, a method that unifies the LUPI framework with the knowledge distillation paradigm. Bengio \etal~\cite{bengio2009curriculum} argued that instead of employing samples at random it is better to present samples organized in a meaningful way so that less complex examples are presented first. Curriculum learning \cite{bengio2009curriculum, tscl, pentina2015curriculum}, which is the learning strategy that implements this paradigm, employs the prior knowledge learned from the first ``easier'' tasks to improve the performance on ``harder'' ones that are learned at a later stage. Such a learning process can exploit prior knowledge to improve subsequent classification tasks but it does not scale up to many tasks since each subsequent task has to be learned individually. 

\noindent\textbf{Adaptive SVM} \cite{yang2007cross}: 
In this section, we provide some theoretical background on Adaptive SVM \cite{yang2007cross} and highlight its differences from a regular SVM, and then we formulate SVM+ \cite{vapnik2009new}, which employs the LUPI paradigm. 
In the standard paradigm of supervised learning for binary classification, the training set consists of \(N\) tuples of feature vectors \(\mathbf{x}_i\), along with their respective labels \(y_i\), represented as \(\{(\mathbf{x}_i, y_i)\}_{i=1}^{N}, \mathbf{x}_i \in \mathbb{R}^d\), where \(d\) is the number of features of each sample and \mbox{\(y_i \in \{-1,+1\}\)}. The standard SVM classifier finds a maximum-margin separating hyperplane between the two classes and solves the following constrained optimization problem: 
\begin{equation}\label{eq:svm}
\small
\begin{aligned}
& \underset{\substack{\textbf{w}, \; b}}{\text{minimize}}
\frac{1}{2}||\mathbf{w}||^2 + C\sum_{i=1}^{N}\xi_i, \\
& \text{s.t.} \; \; y_i\big(\langle \mathbf{w},\mathbf{x}_i\rangle+b\big) \geq 1-\xi_i,  \xi_i\geq 0,
\end{aligned}
\end{equation}
\normalsize
where \(\mathbf{w}\) represents the weight vector, \(||\mathbf{w}||^2\) is the size of the margin, \(b\) is the bias parameter, \(\xi\) is the slack variable for one training sample that indicates the deviation from the margin borders and \(C\) denotes the penalty parameter.

Suppose we are given a training set comprising features \(\mathbfcal{X}^s\) of dimensions \(l_1 \times d\) and binary labels \(\mathbfcal{Y}^s\) of dimensions \(l_1 \times 1\). We will refer to this domain as source domain and we will train a classifier \(f^s(\mathbf{x}_i^s)\) which predicts the respective labels. We are also given another dataset (the target domain) which comprises features \(\mathbfcal{X}\) of dimensions \(l_2 \times d\) and binary labels \(\mathbfcal{Y}\) of dimensions \(l_2 \times 1\). If this dataset had a plethora of data samples then a classifier could be learned on \((\mathbfcal{X},\mathbfcal{Y})\) using Eq.~(\ref{eq:svm}) or any other classification paradigm. However, the target domain might be comprised of mostly unlabeled data, and thus learning from a dataset with a few labeled samples would result in a classifier with high variance on its predictions and poor generalization. Furthermore, if the previously learned classifier \(f^s\) was applied to the new data, then it would yield poor performance, since the target domain might originate from a different distribution. 

To address these challenges, Yang \etal~\cite{yang2007cross} introduced Adaptive SVM. They proposed to adapt an auxiliary classifier to the target domain by learning a ``delta function'' between the decision functions of the auxiliary and target classifiers using an objective function extended from standard SVMs. Intuitively, using an Adaptive SVM is similar to domain adaptation or transferring knowledge between tasks. The adaptation is performed using \(\Delta f(\mathbf{x}) = \mathbf{w}^T \langle \mathbf{x},\mathbf{x}\rangle\) on the basis of \(f^s(\mathbf{x})\), where \(\langle \mathbf{x},\mathbf{x}\rangle\) is a feature map to project each feature vector \(\mathbf{x}\) to a higher dimension via the kernel trick (also referred to as \(K(\mathbf{x},\mathbf{x})\)). Thus, in Adaptive SVM we are interested in learning the function \(f(\mathbf{x}) = f^s(\mathbf{x}) + \Delta f(\mathbf{x}) = f^s(\mathbf{x}) + \mathbf{w}^T \langle \mathbf{x},\mathbf{x}\rangle\). The Adaptive SVM objective function is defined as follows: 
\begin{equation}\label{eq:asvm}
\small
\begin{aligned}
& \underset{\mathbf{w},b}{\text{minimize}}
\frac{1}{2}||\mathbf{w}||^2  + C \sum_{i=1}^{l2} \xi_i \\
& \text{s.t.} \; \; y_i f^s(\mathbf{x}_i) + y_i \mathbf{w}^T \langle \mathbf{x}_i,\mathbf{x}_i\rangle \geq 1 - \xi_i, \; \; \xi_i \geq 0, 
\end{aligned}
\end{equation}
\normalsize
where \(\mathbf{w}\) refers to the parameters of \(\Delta f(\mathbf{x})\) and thus, \(||\mathbf{w}||^2 = ||f-f^s||^2\). This implies that Adaptive SVM seeks to minimize the distance between the adapted decision and the decision of the classifier \cite{yang2007cross} in the source domain. Using the computed support vectors, we obtain the adapted decision function in which a new testing sample of the primary dataset is first passed through the decision function of the source domain classifier and then from the adapted decision function. A parameter (\(\Gamma\)) that controls the weight of the decision of the auxiliary classifier can be added in Eq.~(\ref{eq:asvm}) as in the method of Aytar and Zisserman \cite{aytar2011tabula}. To avoid adding an extra parameter that also needs to be cross-validated we will refrain from using it in the rest of our paper. 

\noindent\textbf{Learning Using Privileged Information (SVM+)} \cite{vapnik2009new}: In the LUPI setup, during the training phase, instead of tuples of features and labels we are given triplets \(\{(\mathbf{x}_i,\mathbf{x}_i^*,y_i)\}{_{i=1}^{N}}, \mathbf{x}\in\mathbb{R}^d,\) \(  \mathbf{x}^*\in\mathbb{R}^{d^{*}},\) \mbox{\(y_i \in \{-1,+1\}\)}, where feature vectors \(\mathbf{x}^*\) represent the additional (\ie, privileged) information. During the testing phase, features from the privileged space \(\mathbfcal{X^*}\) are not available. The goal of LUPI is to exploit the privileged information during the training phase to learn a model that further constrains the solution in the original space \(\mathbfcal{X}\), and thus it can more accurately describe the testing data. In this paradigm, the slack variables \(\xi_i\) are parameterized as a function of privileged information \(\xi_i(\mathbf{w}^*,b^*) = \langle \mathbf{w}^*,\mathbf{x}_i^* \rangle + b^*\). The SVM+ algorithm, which implements LUPI in the training phase, solves the following minimization problem: 
\begin{equation}\label{eq:svmplus}
\small
\begin{aligned}
&\underset{\substack{\mathbf{w},\,b,\,\mathbf{w}^*,b^*}}{\text{minimize}}
\; \frac{1}{2}\big(||\mathbf{w}||^2 +\gamma||\mathbf{w}^*||^2\big) + C\sum_{i=1}^{N}\xi_i(\mathbf{w}^*,b^*), \\
&\text{s.t.}
\; y_i\big(\langle \mathbf{w},\mathbf{x}_i \rangle +b\big) \geq 1-\xi_i(\mathbf{w}^*,b^*), \; \; \xi_i(\mathbf{w}^*,b^*) \geq 0,
\end{aligned}
\end{equation}
\normalsize
where \(\gamma\) controls the weight of the privileged information in the correcting (\ie, privileged) space. In SVM+ the decision function \(f(\mathbf{x})\) is the same with SVM, as at test time the privileged information is not available. For additional information regarding the dual formulations of each of the objective functions, the interested reader is encouraged to refer to the original publications \cite{vapnik2009new, yang2007cross} and for fast algorithms for both the linear and the kernel cases to the work of Li \etal~\cite{Li_2016_CVPR}.

\section{Adaptive SVM+}\label{sec:asvm}
We introduce Adaptive SVM+, a novel method to perform domain transfer using privileged information. In Adaptive SVM+ one is provided with two sets of data that might be originating from completely different distributions. A classifier is first learned in the source domain which may also have additional information. Since the data in the target domain contain privileged information, a new objective function is needed based on SVM+ which at the same time minimizes the distance between the adapted decision function \(f(\mathbf{x})\) (computed on the triplets \((\mathbfcal{X,X^*},\mathbfcal{Y})\) of the target domain) and the auxiliary function \(f^s(\mathbf{x})\) obtained from the decision function in the source domain. 

\noindent\textbf{Objective Function}: Adaptive SVM+ seeks to minimize the distance of the data in the target and source domains only in the original space and not in the privileged. The reason for this is twofold. First, if we sought to minimize the distance between the privileged information of two different domains, we would make a strong assumption that would not hold in most cases. Second, since such information is not available at test time, if we minimized the distance between the privileged information of the two domains, we would have to leverage information learned in the privileged space of the source domain in the new target domain, which would break  the intuition behind learning with an intelligent teacher as in LUPI. Thus, the new objective function of Adaptive SVM+ is defined as follows:   
\begin{equation}\label{eq:asvmpluslambda}
\small
\begin{aligned}
& \underset{\mathbf{w},b, \mathbf{w}^*, b^*}{\text{minimize}} \; \frac{1}{2}\big(||\mathbf{w}||^2 + \gamma||\mathbf{w}^*||^2\big) + C \sum_{i=1}^{l} \big(\langle \mathbf{w}^*,\mathbf{x}^*_i \rangle +b^*\big)\\
& \text{s.t.}  \;\;\; y_i f^s(\mathbf{x}_i) + y_i (\langle \mathbf{w},\mathbf{x}_i \rangle +b) \geq 1 - \big(\langle \mathbf{w}^*,\mathbf{x}^*_i \rangle +b^*\big), \\ 
& \;\;\;\;\;\;\big(\langle \mathbf{w}^*,\mathbf{x}_i^* \rangle +b^*\big) \geq 0,
\end{aligned}
\end{equation}
\normalsize
To solve this problem, we construct the (primal) Lagrangian:
\begin{equation}\label{eq:asvmpluslag}
\footnotesize
\begin{aligned}
& L(\mathbf{w},b,\mathbf{w}^*,b^*, \mathbf{\alpha}, \mathbf{\beta}) = \frac{1}{2}\big(||\mathbf{w}||^2 + \gamma||\mathbf{w}^*||^2\big) + \\
& + C \sum_{i=1}^{l} \big(\langle \mathbf{w}^*,\mathbf{x}_i^* \rangle +b^*\big) -\sum_{i=1}^{l} \beta_i\big(\langle \mathbf{w}^*,\mathbf{x}_i^* \rangle +b^*\big) \\
& -\sum_{i=1}^{l} \alpha_i\Big( y_i f^s(\mathbf{x}_i) + y_i (\langle \mathbf{w},\mathbf{x}_i \rangle +b) - \big(1 - (\langle \mathbf{w}^*,\mathbf{x}_i^* \rangle +b^*)\big) \Big),
\end{aligned}
\end{equation}
\normalsize
where \(\alpha, \beta \geq 0\) are the Lagrange multipliers. The dual formulation of the problem is defined as follows: 
\begin{equation}\label{eq:asvmplusdual}
\small
\begin{aligned}
& \underset{\alpha,\beta}{\text{minimize}} \; \; \frac{1}{2}\sum_{i,j=1}^{l} \alpha_i \alpha_j y_i y_j K(x_i,x_j) - \sum_{i=1}^{l} (1-\lambda_i)\alpha_i + \\ 
& + \frac{1}{2\gamma}\sum_{i,j=1}^{l} (\alpha_i + \beta_i - C)(\alpha_j + \beta_j - C) K^*(x_i^*,x_j^*)  \\ 
& \;\;\;\;\; \text{s.t.} \sum_{i=1}^{l} (\alpha_i + \beta_i - C)  = 0, \;\; 0\leq\alpha_i, \beta_i,
\end{aligned}
\end{equation}
\normalsize
where similar to Adaptive SVM, \(\lambda_i = y_i f^s(\mathbf{x}_i)\) and \(K^*\) is the kernel in the privileged space. Minimizing Eq.~(\ref{eq:asvmplusdual}) over \(\alpha, \beta\) is a quadratic programming problem, which provides the support vectors of the primary data \(\mathbf{x}_i\) in the original space \(\hat{\alpha}\) and in the privileged space \(\hat{\beta}\), which can be used for the correcting function. At testing time, only primary tuples \(\mathbfcal{X},\mathbfcal{Y}\) are available and privileged information \(\mathbfcal{X^*}\) is absent. Thus the decision function of Adaptive SVM+ is no different than that of Adaptive SVM, which is defined as:
\begin{equation}\label{eq:aSVMdecMore}
\small
\begin{aligned}
f(\mathbf{x}) &= f^s(\mathbf{x}) + \sum_{i = 1}^{l_2} y_i\hat{\alpha}_i K(x_i,\mathbf{x}) + b \\
&= \sum_{i = 1}^{l_2} y^s_i\alpha^s_i K(x^s_i,\mathbf{x}) + b^s + \sum_{i = 1}^{l_2} y_i\hat{\alpha}_i K(x_i,\mathbf{x}) + b ,
\end{aligned}
\end{equation}
\normalsize
\begin{algorithm}[t]
	\footnotesize
	\SetKwInOut{Input}{Input}
	\SetKwInOut{Output}{Output}
	\Input{Training triplets (\(\mathbfcal{X, X^*},\mathbfcal{Y}\)), testing features \(\mathbfcal{X}_t\), decision function in the source domain \(f^s(\mathbf{x})\) }
	\(\hat{\alpha}, \hat{\beta} \leftarrow \) compute support vectors using the triplets \(\mathbfcal{X, X^*},\mathbfcal{Y}\) by minimizing Eq. (\ref{eq:asvmplusdual})\\
	\(f(\mathbf{x}) \leftarrow\) construct the decision function using the obtained support vectors \(\hat{\alpha}\), testing features \(\mathbfcal{X}_t\) and  Eq. (\ref{eq:aSVMdecMore})\\
	\(Y_p\leftarrow\) obtain predictions by computing the sign of \(f(\mathbf{x})\)\\
	\Output{Class Predictions in the target domain \(Y_p\)}
	\caption{Adaptive SVM+}
	\label{alg1} 
\end{algorithm}

\begin{figure}[t] 
	\centering
	\includegraphics[width=0.49\textwidth]{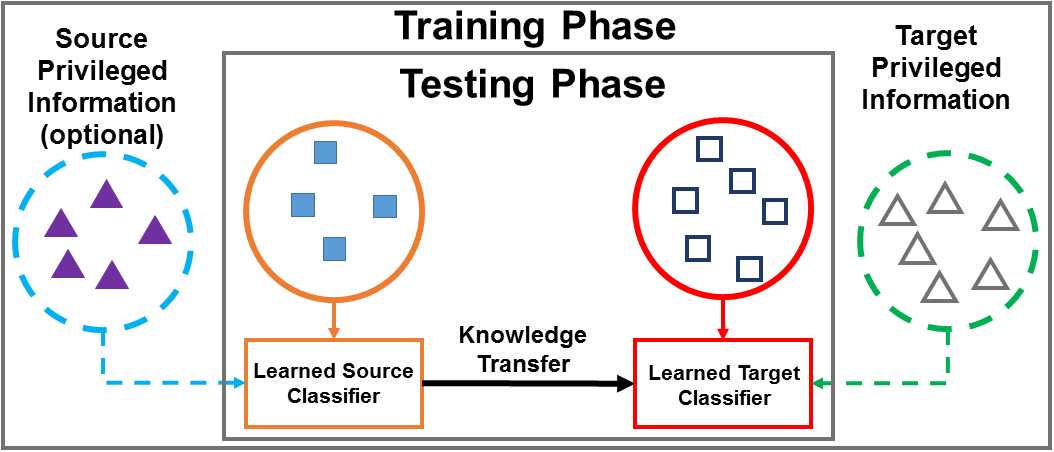}
	\caption{Training and testing phases of Adaptive SVM+. A classifier is first learned in the source domain, which might or might not have privileged information (dashed blue circle). This domain is then adapted to the target domain in which the Adaptive SVM+ classifier, which uses additional features at training time, is learned.} 
	\label{fig:LearningScheme}
\end{figure}

\noindent\textbf{Key Characteristics and Differences}: Adaptive SVM+, described in Algorithm~\ref{alg1}, takes as an input features in the original space \(\mathbfcal{X}\), privileged features \(\mathbfcal{X^*}\), and labels \(\mathbfcal{Y}\), as well as the decision function \(f^s\) learned in the source domain. Learning \(f^s\) is not constrained to a specific classifier (Naive Bayes, SVMs and decision trees are all valid options \cite{yang2007cross}) or to a specific learning paradigm since privileged information can also be exploited during the learning stage of \(f^s\). Using the features in the new domain (depicted with a red circle in Figure~\ref{fig:LearningScheme}) the proposed paradigm aspires to minimize the distance between the two domains in the original space, while at the same time utilizing privileged information to learn a better decision function in the original space of the target domain. The differences in the dual formulation between SVM+ and Adaptive SVM+ correspond to the introduction of an extra term in Eq.~(\ref{eq:asvmplusdual}), which incorporates information from the source domain, and the lack of an additional constraint that exists in SVM+, but does not exist in our proposed learning scheme. This constraint is embedded in the objective function and is also absent in the Adaptive SVM formulation compared to the regular SVM. 


\section{Experiments}\label{sec:exper}
To verify the effectiveness of our method, we conducted evaluations and report state-of-the-art results in two datasets: Animals with Attributes \cite{lampert2014attribute} and INTERACT \cite{antol2014zero}. 

\subsection{Animals with Attributes Dataset}
We followed the same experimental procedure with \cite{motiian2016information, sharmanska2013learning, wang2015classifier} in which out of the 50 animal classes, we used the 10 testing classes for a total of 6,180 images. 

\noindent\textbf{Features}: For the primary space, we used the same set of features with \cite{motiian2016information, sharmanska2013learning}, which are \(L_1\)-normalized 2,000 dimensional SURF descriptors, provided along with the dataset. For the privileged space, we computed attribute predictions for each of the 85 attributes using the DAP model \cite{lampert2014attribute}. Sharmanska \etal~\cite{sharmanska2014learning} collected Mechanical Turk annotation of images to define easy and hard samples for eight out of the 10 classes of the AwA dataset. The scores are in the range from 1 (hardest) through 16 (easiest), which are linearly scaled to \([0,2]\), where values less than or equal to one correspond to hard samples and scores greater than one to easy samples. We use these scores to define our source and target domains as we first learn the former (the easy samples) and then the latter (the hard samples) by performing domain adaptation at the same time. 

\noindent\textbf{Evaluation Metric}: We evaluate Adaptive SVM+ by reporting average precision (AP) results, which correspond to the area under the precision-recall curve. The train/test split is repeated 20 times and average AP results along with standard error over all possible configurations are reported.

\noindent\textbf{Model Selection}: The same joint cross validation scheme with \cite{motiian2016information, sharmanska2013learning} is used, during which the best parameters are selected based on 5-fold cross validation and are then used to re-train the complete training set. In both the source and the target domains the parameter C and the parameter \(\gamma\), which controls the influence of the privileged space, are searched within \(\{10^{-4},\dots, 10^4\}\). 

\noindent\textbf{Training}: We train 45 binary classifiers for each class pair combination (\eg, chimpanzee versus giant panda) using 50 and 200 samples per class for training and testing, respectively. We first train an SVM+ classifier on the easy samples (\ie, source domain) and then an Adaptive SVM+ classifier on the hard samples. When no easy/hard scores are available for one of the two classes, we report SVM+ classification results without domain transfer. To perform a fair comparison with the rest of the methods: (i) a linear kernel is used in all domains and both original and privileged spaces; and (ii) the easy/hard ratio is preserved in the reported results, which means that if in one class, 75\% of the samples are easy and the rest are hard, after we train both classifiers we randomly select 75\% of the easy predictions and 25\% of the hard predictions to report the final AP results.

\begin{table}[t]
	\centering
	\caption{Average precision (AP) results on the Animals with Attributes dataset, with attributes as privileged information and easy/hard sample annotation as source/target domains. On the left side, we provide complete results in both domains for a fair comparison. On the right side, we provide as reference the performance of the respective methods in each domain separately.}
	\label{tab:AwATable}
	\footnotesize
	\begin{tabular}[t]{lc}
		\toprule
		Method & AP \\
		\midrule
		SVM  & 87.32 \\
		SVM+ \cite{vapnik2009new} & 87.58 \\
		Adaptive SVM \cite{yang2007cross}  & 87.94  \\
		RankTr \cite{sharmanska2013learning} & 87.93   \\
		LIR  \cite{wang2015classifier}   & 88.13 \\
		LMIBPI \cite{motiian2016information} & 88.38\\
		\textbf{Adaptive SVM+}  &    \textbf{88.66}  \\
		\bottomrule
	\end{tabular}
	~
	\begin{tabular}[t]{lc}
		\toprule
		Method - Domain & AP \\
		\midrule
		SVM (easy) & 89.93 \\
		SVM+ (easy)     &  90.10      \\
		\midrule
		SVM (hard) & 78.17 \\
		Adaptive SVM (hard) & 79.63 \\
		SVM+ (hard) &  78.78     \\
		\textbf{Adaptive SVM+} (hard) &   80.10    \\
		\bottomrule
	\end{tabular}
\end{table} 

\noindent\textbf{Discussion of Results}: We provide a summary of the mean AP results for all tasks in Table~\ref{tab:AwATable}. Using the exact same features and evaluation protocol, our method achieves state-of-the-art results. Adaptive SVM+ is better than the rest in 21 out of 45 tasks, 13 of which are statistically significant over the second best method (z-test). For the rest of the methods, LMIBPI \cite{motiian2016information} achieved higher AP 15 times, RankTr \cite{sharmanska2013learning} 5, and LIR  \cite{wang2015classifier} 4 times. On the right side of Table~\ref{tab:AwATable}, we provide domain specific results along with the method from which we observe that: (i) privileged information is beneficial, as both SVM+ and Adaptive SVM+ perform better than their counterparts; and (ii) domain adaptation is beneficial, as in both the Adaptive SVM and Adaptive SVM+ cases in the target domain there is a performance increase. 

\subsection{INTERACT Dataset}\label{ssec:interact}
The INTERACT dataset \cite{antol2014zero} comprises 3,172 images of 60 fine-grained categories of interactions between two people such as ``laughing with'', ``is lying in front of'', or ``is walking after''. Additionally, illustrations in the form of clip art are provided depicting the same 60 fine-grained categories in two different level settings: (i) category-level in which images and illustrations are collected independently, and (ii) instance-level in which 2-3 illustrations of the same interaction category are collected for a given image. We followed the same experimental procedure with the method of Sharmanska and Quadrianto \cite{sharmanska2016learning} for the instance-level setting. They proposed a framework called SVM MMD to ``learn from the mistakes of others'' by minimizing the distribution mismatch between errors made in images and in privileged data (\ie, illustrations) using the Maximum Mean Discrepancy (MMD) criterion. Adding a regularizer, based on the MMD criterion to reduce the data distribution mismatch in a LUPI setup was initially introduced by Li \etal~\cite{li2014exploiting} to perform image categorization and retrieval. 

\noindent\textbf{Features}: Both real images and illustrations are represented by a 765-dimensional feature vector capturing human pose information, expressions, relation (from person 1 to person 2) and appearance and are provided with the dataset. As in \cite{sharmanska2016learning} we pair each real image with a randomly selected (among the two or three) illustration per image. Clip art illustrations are used as a source domain and real images as a target domain. 

\noindent\textbf{Evaluation Metric}: To evaluate our approach, we used classification accuracy. The train/test split process is repeated 20 times and average results along with standard error across repeats are reported. 

\noindent\textbf{Model Selection}: Following the evaluation protocol of Sharmanska and Quadrianto \cite{sharmanska2016learning}, we select the parameter C from \(\{10^0,\dots, 10^5\}\) and in the privileged space the values for both C and \(\gamma\) are obtained from \(\{10^{-4},\dots, 10^4\}\). Once the parameters are obtained using a 3-fold cross-validation scheme, we use them to re-train the complete set. 

\noindent\textbf{Training}: We trained 60 one-versus-rest binary classifiers to predict the interaction of interest against the rest of the interactions. Similar to \cite{sharmanska2016learning}, we trained Adaptive SVM+ using linear kernels and by sampling 25 positive vs 25 negative images.  For testing, we use the remaining positive images balanced with negative samples. Privileged features comprise a randomly selected instance-level clip art illustration, which depicts two sketches of humans imitating the same interaction. Contrary to the AwA dataset, the decision function in the source domain is learned without privileged information as we simply train an SVM on clip art illustrations. At testing time, Adaptive SVM+ is presented only with real images of interactions of humans and no information related to the clip art illustrations is available. 

\noindent\textbf{Discussion of Results}: A summary of the classifications results on the INTERACT dataset is presented in Table~\ref{tab:interactTable}. When using only linear kernel, our method performs better than the state of the art. Although the improvement can be seen as marginal in the linear kernel case, note that all methods are marginally better than a regular SVM, since some interactions are very similar to some others (\eg, walking to, walking away from, walking with), which makes the accurate classification of such tasks very challenging. Adaptive SVM+ is more accurate in 32 out of the 60 interactions, SVM MMD \cite{sharmanska2016learning} in 19, and the rest are attributed to SVM, SVM+ and Adaptive SVM. When RBF kernels are used, there is a 2.81\% relative improvement. 

\begin{table}[t]
	\centering
	\caption{Classification accuracy results on the INTERACT dataset, with one instance-level clip art per sample as privileged information and illustrations/real images as source/target domains.}
	\label{tab:interactTable}
	\footnotesize
	\begin{tabular}{lc|lc}
		\toprule
		Method & Cl. Acc. & Method & Cl. Acc. \\
		\midrule
		SVM Images & 80.51 & SVM+ \cite{vapnik2009new} & 80.93 \\
		SVM Illustrations  & 77.32 & SVM MMD \cite{sharmanska2016learning} & 81.58 \\
		SVM Combined  & 79.91 & \textbf{Adaptive SVM+} & \textbf{81.87} \\
		Adaptive SVM  & 80.22 & \textbf{Adaptive SVM+ (RBF)} & \textbf{83.87} \\
		\bottomrule
	\end{tabular}
\end{table}

\subsection{In the Deep Learning Era is Privileged Information Necessary?} 
\begin{figure}[t] 
	\centering
	\includegraphics[width=0.48\textwidth]{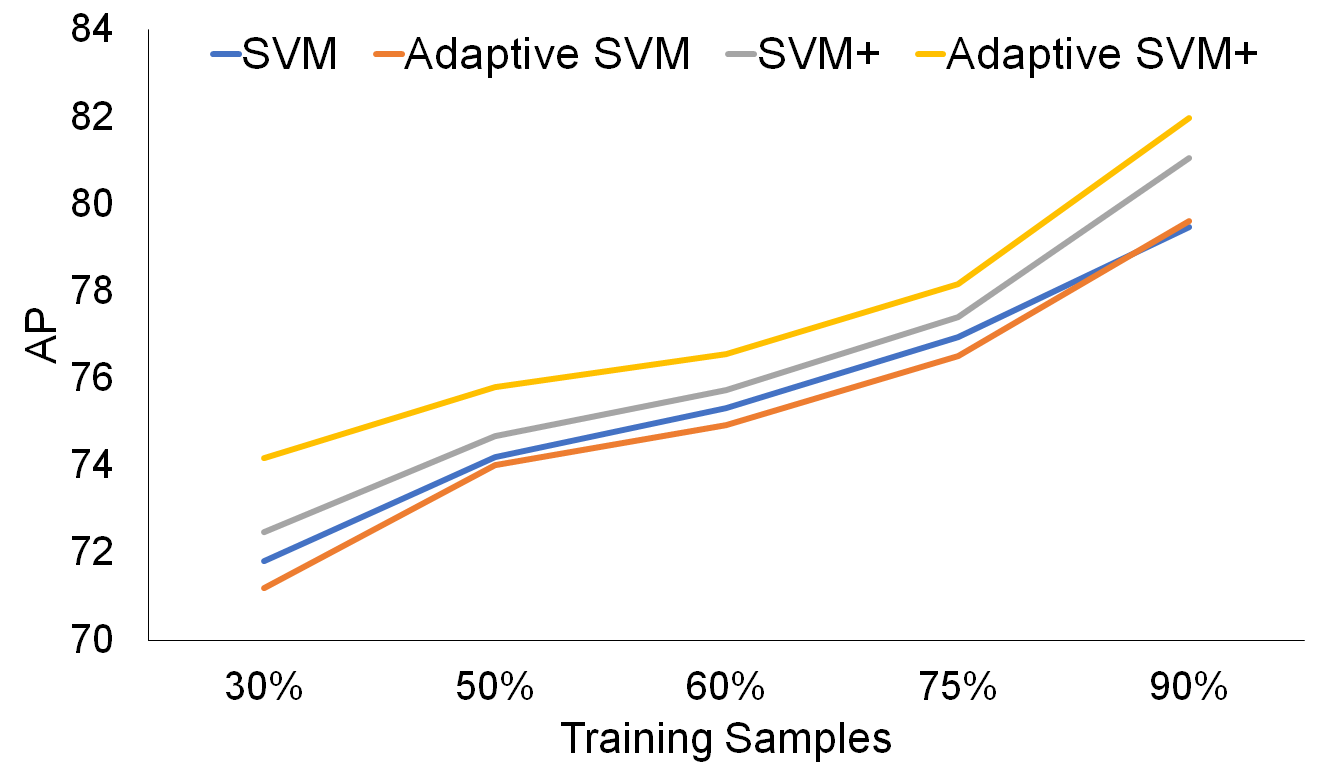}
	\caption{Average precision of different classifiers using VGG features from the INTERACT dataset versus the ratio of training samples over the whole set of features.} 
	\label{fig:vgg_inter}
\end{figure}

Interested in evaluating our proposed approach with ConvNet-based features, we first trained as a baseline an SVM on the Animals with Attributes dataset with features extracted from the last fully-connected layer of the VGG network \cite{simonyan2014very}. We observed that the AP over all tasks was over 99\%, which is reasonable since ImageNet comprises more than a hundred different classes of animals and thus, the extracted feature representations are very discriminative for such a task. However, for the INTERACT dataset, which contains human interactions (that are not part of the ImageNet classes), the obtained results did not reach the same performance. Using z-score normalized VGG features, linear kernels and the same hyper-parameters with Section~\ref{ssec:interact} we trained all four classifiers (\ie, SVM, Adaptive SVM, SVM+ and Adaptive SVM+) on different ratios of training samples over the whole feature set. The average precision for the different models with respect to the ratio of training samples is depicted in Figure~\ref{fig:vgg_inter}. We observed that when training samples constitute 75\% or more of the whole dataset, privileged information can be beneficial as for both SVM+ and Adaptive SVM+ there is an increase on the average precision. Note that there are approximately 60 samples for each of the positive and negative classes which explains why the performance is not higher. The aim of our proposed approach and the rest of the literature, was not to achieve the best results possible on these datasets, but under the same evaluation protocol to investigate to what extent privileged information and domain adaptation can be beneficial. 

However, an interesting discussion arises from these results. Since representation learning with ConvNets is a very powerful feature extractor, is privileged information necessary? We believe that the answer to this question is positive for two different reasons. First, there are plenty of challenging benchmarks (\eg, MS COCO \cite{lin2014microsoft}) in which state-of-the-art deep learning techniques have not yet achieved ImageNet-level results. Even on ImageNet, which has been thoroughly benchmarked, a recent work of Chen \etal~\cite{chen2017training} demonstrated that by using segmentation annotations as privileged information better performance may be achieved. Second, there are applications in which annotated data are rare, difficult or even expensive to obtain (\eg, medical data) and pre-trained deep learning models are still not available. In such cases, privileged information in the form of additional features or in the form of domain adaptation \cite{csurka2017domain, ge2017borrowing, tzeng2017adversarial} is still very relevant. 

\section{Conclusion}\label{sec:conc}
Can we leverage privileged information in a domain adaptation setup? Is there a need to exploit such information from a source domain in addition to the privileged information in the target domain? Can we do better than state-of-the-art techniques? In this work, we sought to give an answer to these questions by proposing Adaptive SVM+; a novel yet simple learning paradigm. It combines the advantages of both SVM+ and Adaptive SVM, and seeks to minimize the distance between a source and a target domain while at the same time, utilizing privileged information on the latter. We tested the proposed learning scheme in object recognition and human interaction classification tasks with visual attributes along with human annotations of easy/hard scores and clip-art illustrations of interactions, respectively. We observed that Adaptive SVM+ achieved state-of-the-art results across the board without adding any complexity or extra parameters compared to the available methods. 
	
\subsection*{Acknowledgments}
\noindent{This work has been funded by the UH Hugh Roy and Lillie Cranz Cullen Endowment Fund. All statements of fact, opinion or conclusions contained herein are those of the authors and should not be construed as representing the official views or policies of the sponsors.}
\clearpage

	{\small
		\bibliographystyle{ieee}
		\bibliography{egbib-Copy}
	}
	\clearpage

\end{document}